\documentclass{article}

\usepackage[numbers]{natbib}

\usepackage[utf8]{inputenc} 
\usepackage[T1]{fontenc}    
\usepackage{hyperref}       
\usepackage{url}            
\usepackage{booktabs}       
\usepackage{amsfonts}       
\usepackage{nicefrac}       
\usepackage{microtype}      
\usepackage{xcolor}         
\usepackage{graphicx}

\title{Hashmarks: Privacy-Preserving Benchmarks for High-Stakes AI Evaluation}

\author{%
  Paul Bricman \\
  Straumli AI\\
  \texttt{paul@straumli.ai} \\
}

\begin{document}

\date{December 1, 2023}

\maketitle

\begin{abstract}
  There is a growing need to gain insight into language model capabilities that relate to sensitive topics, such as bioterrorism or cyberwarfare. However, traditional open source benchmarks are not fit for the task, due to the associated practice of publishing the correct answers in human-readable form. At the same time, enforcing mandatory closed-quarters evaluations might stifle development and erode trust. In this context, we propose \textit{hashmarking}, a protocol for evaluating language models in the open without having to disclose the correct answers. In its simplest form, a hashmark is a benchmark whose reference solutions have been cryptographically hashed prior to publication. Following an overview of the proposed evaluation protocol, we go on to assess its resilience against traditional attack vectors (e.g. rainbow table attacks), as well as against failure modes unique to increasingly capable generative models.
\end{abstract}

\section{Introduction}

\bibliographystyle{unsrtnat}

\subsection{Background \& Motivation}

Traditional question-answering (QA) benchmarks have played a crucial role in shaping the trajectory of AI development. They provide standardized metrics that facilitate fair comparisons across research groups and measure progress within the field in a reproducible way \cite{liang_holistic_2023}. For instance, capabilities related to mathematics can generally be gauged by assessing model performance in producing or selecting correct answers for exam questions related to mathematics \cite{hendrycks_measuring_2021, saxton_analysing_2019}. Similar "AI exams" have been used to assess model performance on topics ranging from STEM \cite{hendrycks_measuring_2021-1, auer_sciqa_2023, jin_pubmedqa_2019, clark_think_2018} to humanities \cite{hendrycks_measuring_2021-1, hendrycks_aligning_2023}. In fact, prior work has highlighted the possibility of framing the vast majority of established natural language processing tasks as question-answering tasks \cite{khashabi_unifiedqa_2020}.

Traditional QA benchmarks are typically sourced from crowd-workers, members of the group developing the benchmark, or a mix of the two \cite{liang_holistic_2023, bowman_what_2021}. They typically contain a large number of data points representing individual exercises. Each data point, in turn, contains one question and one or several correct answers. Occasionally, a data point may also contain some relevant context in the form of a separate string, although the context can always be functionally merged with the question without loss of generality \cite{shaham_scrolls_2022, pang_quality_2022}. In addition, when structured as a multiple-choice question, a data point also contains several distractor answers \cite{hendrycks_measuring_2021-1, hendrycks_aligning_2023}. These data points are then aggregated into one or more standardized files, and are finally made public for other groups to easily make use of them in evaluating language models \cite{liang_holistic_2023, noauthor__nodate}.

While traditional QA benchmarks have been instrumental in assessing models across a wide range of benign domains, there is a growing need to better understand capabilities that relate to sensitive topics, such as bioterrorism or cyberwarfare \cite{shevlane_model_2023}. Greater insight into, for instance, AI-enabled biorisk, could help calibrate the stringency of regulations and policies in a way that is proportional to the associated threat. Underestimating the level of risk could, among others, enable bad actors to more easily weaponize this technology as a means of causing harm at scale. Conversely, overestimating the level of risk posed by a given class of generative models could also reduce the upsides being captured through beneficial usecases.

Naively applying traditional QA benchmarks on such sensitive topics, however, would face a major obstacle. By disclosing the reference solutions of the exam questions included in a hypothetical traditional benchmark on bioterrorism, one would inadvertently publish a veritable compendium on a topic better left undiscussed. The collection of questions coupled with correct answers would, in essence, provide a publicly-available "FAQ" on knowledge related to the topic. Needless to say, more secure ways of assessing hazardous capabilities in generative models are required.

\subsection{Related Work}

Fortunately, the field of cryptography is rich with ideas and practices that enable parties to prove statements to each other, such as the fact that a certain answer is incorrect, without disclosing any other sensitive information, such as the correct answer. In password authentication, for instance, a server can determine whether or not a candidate password corresponds to the correct password without actually having knowledge of the correct password \cite{noauthor_password_nodate}. This is typically achieved by first irreversibly "hashing" the password during user registration using a cryptographic hash function \cite{noauthor_password_nodate, naor_universal_1989, noauthor_argon2_nodate}. Then, during user authentication, the server checks whether the hashed version of the candidate password matches the hashed version of the correct password. Not requiring knowledge of the correct password in cleartext during authentication means that attackers gaining access to the server cannot simply steal user passwords \cite{noauthor_password_nodate}. However, there are additional complexities involved in this technology which we will turn to when discussing the resilience of the proposed protocol against a range of attacks.

Another fertile field for developing secure evaluation protocols is federated learning \cite{li_federated_2020}, a decentralized machine learning paradigm that enables parties to collaboratively train models without sharing their raw training data. This is achieved by, for instance, sharing and aggregating \textit{models} trained on local data, rather than the training datasets themselves \cite{ma_layer-wised_2022}. Differential privacy, as another example, largely focuses on superficially corrupting the data being aggregated by adding noise, in order to protect individual privacy while still extracting meaningful insights \cite{dwork_calibrating_2006}. These techniques ensure that the learning process is privacy-preserving.

Drawing on these cryptographic and federated learning concepts, the following section describes a privacy-preserving evaluation protocol that aims to assess the capabilities of language models in sensitive domains without exposing unnecessary details about the correct answers.

\section{Method}

\subsection{Problem Statement}

Before documenting the protocol itself, it is instructive to explicitly state the problem being addressed. First, there are several experts who all possess knowledge related to a given dual-use research direction. Second, there is an auditor who wishes to package and share the experts' knowledge with third-parties in a way that would, as much as possible, (1) allow third-parties to verify their existing level of knowledge on the topic, while (2) preventing third-parties from acquiring more knowledge on the topic than they previously had.

\subsection{Hashmarking Protocol}

\begin{figure}
  \centering
  \frame{\includegraphics[width=\textwidth]{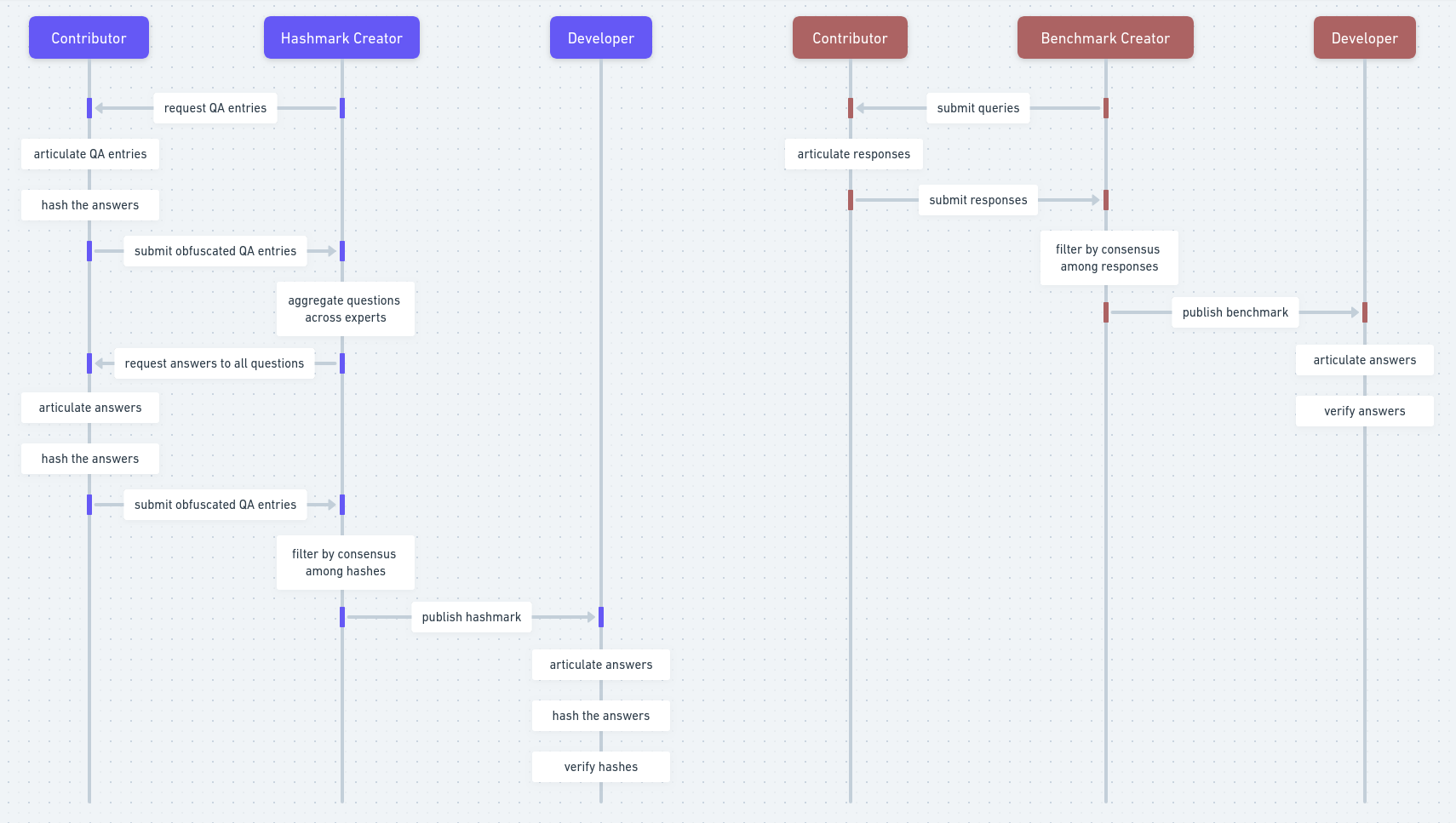}}
  \caption{Side-by-side sequence diagrams for hashmarking (left) and traditional benchmarking (right). Notably, the hashmark creator does not gain access to the human-readable reference solutions, only to their hashed versions. In addition, developers can only uncover the reference solutions behind a hashmark if they already possess them.}
  \label{fig:protocol}
\end{figure}

In the context of the problem statement articulated above, we now describe hashmarking, a simple protocol that would enable the auditor to share expert knowledge in a way that enables knowledge verification while preventing knowledge acquisition pertaining to dual-use research (Fig. \ref{fig:protocol}).

First, each expert articulates a series of question-answer pairs that relate to their topic of expertise. Then, they all hash the correct answers using a slow hashing algorithm \cite{noauthor_argon2_nodate}. In addition, they all use the associated questions as salt \cite{noauthor_password_nodate} in the process of hashing the correct answers. Once the hashing of the correct answers is complete, the experts send their obfuscated question-answer pairs to the auditor.

Second, the auditor sends each individual expert all the cleartext questions contributed by all \textit{other} experts, without the associated answers in hashed form. Then, each expert attempts to provide an answer to each of the questions provided by the other experts. If not confident in the answer to such a question -- there may be fields relating to a dual-use research direction that exhibit limited overlap -- experts attach an empty answer. After answering this second round of questions, the experts process their new answers in the same exact way as before. They hash them using the questions as salt, package them together, and send the result to the auditor.

Third, the auditor discards those question-answer pairs that have less than a threshold number of non-empty answers. Then, the auditor also discards those question-answer pairs that do not exhibit consensus among the hashed answers contributed by the various experts. By filtering for inter-annotator agreement, the auditor attempts to improve the quality of the question-answer collection contributed by the experts without possessing the cleartext correct answers themselves.

Fourth, the auditor publishes the filtered collection of cleartext questions and hashed answers in the open. Third-parties are now able to quantify their knowledge on the topic by attempting to answer the questions themselves, hashing them exactly as the experts have done, and checking whether the resulting hashes correspond to the hashes of the correct answers. At the same time, third-parties lacking the expert knowledge in the first place have a much harder time deriving value from the otherwise public hashmark.

\subsection{Desiderata for Hashmark Entries}

While the hashmark protocol provides valuable security benefits by hindering dual-use knowledge acquisition while still enabling verification, it also has limitations. The primary constraint comes from the fact that even answers that differ by a handful of characters are hashed in completely different ways, due to the nature of cryptographic hash functions \cite{noauthor_password_nodate, naor_universal_1989, noauthor_argon2_nodate}. This means that questions included in a hashmark need to have a narrow, well-defined, unambiguous answer. For instance, a question could ask for the standard name of a certain chemical. Similarly, a question could ask for the simplified molecular-input line-entry system (SMILES) representation \cite{weininger_smiles_1988} of the same chemical.

The second constraint comes from the fact that, while it is virtually impossible to evaluate the inverse of a secure cryptographic hash function for a given hash, one could still uncover the cleartext solutions by brute-forcing a large number of candidate answers until obtaining a match \cite{noauthor_argon2_nodate, bosnjak_brute-force_2018}. This means that questions included in a hashmark should preferably call for obscure, non-trivial answers. If, instead, a question called for a "yes" or "no" answer, then it could be trivially attacked by hashing both possible answers and identifying the matching hash. Besides encouraging experts to take into account this second desideratum, there are a number of design choices and hyperparameters that can be employed to structurally mitigate related attacks, a topic we turn to in the next section.

In sum, questions included in a hashmark should call for answers that are obscure, yet unambiguous. 

\section{Security}

\subsection{Against Traditional Attacks}

\subsubsection{Brute-Force \& Dictionary Attacks}

The security practices involved in password storage have evolved to mitigate emerging classes of attacks. First, a naive approach to password storage might be to store them in cleartext. However, that would trivially grant attackers access to the actual passwords in the case of a breach. Second, a more advanced approach to password storage might be to encrypt them in a reversible way. However, an attacker might be able to locate the decryption key on the server and so gain access to the passwords by decrypting them.

Third, an even more advanced approach to password storage might be to hash them in an irreversible way \cite{noauthor_password_nodate}. However, an attacker might still be able to carry out a brute-force attack by hashing each and every possible sequence of characters until identifying the one that hashes to the same hash as the original \cite{bosnjak_brute-force_2018, alkhwaja_password_2023}. Similarly, an attacker might also be able to carry out a dictionary attack by hashing a manually-curated list of candidate passwords until potentially identifying one that matches the original hash \cite{bosnjak_brute-force_2018, alkhwaja_password_2023}.

In response to brute-force and dictionary attacks, \textit{slow hashing} has been developed through a class of cryptographic hash functions that are intentionally compute-intensive and/or memory-intensive to evaluate. Slow hashing algorithms include bcrypt \cite{provos_future-adaptable_nodate}, scrypt \cite{percival_stronger_nodate}, and argon2 \cite{noauthor_argon2_nodate}. All of these algorithms can be configured to induce a particular computational and memory cost by adjusting several parameters. In the context of a user-facing authentication system, developers might opt for the slowest possible hashing that is still admissible in terms of user experience. Towards one extreme, the fastest available hashing might enable the fastest authentication times for users, yet might also enable attackers to churn through candidate passwords in a short amount of time through brute-force and dictionary attacks. Towards the other extreme, the slowest available hashing might induce an unpleasantly slow authentication experience for users, yet might be more effective in preventing these types of attacks.

The way brute-force and dictionary attacks would translate to the hashmark protocol is as follows. After downloading a public hashmark, attackers might attempt to run brute-force or dictionary attacks in an attempt to uncover the cleartext answers to the questions related to dual-use research. Brute-force attacks would similarly involve trying each and every possible sequence of characters as a candidate answer, while dictionary attacks would involve sourcing a list of possible answers to a given question (e.g. a list of the standard names of common chemicals).

Fortunately, similar to how slow hashing can help mitigate password cracking, it can also help mitigate the uncovering of the reference solutions in a hashmark. In addition, it might be feasible to prioritize security over the "user experience" of contributing experts and model evaluators. In other words, it might be feasible to acclimatize to mildly inconvenient hashmark creation and verification times, in order to render brute-force and dictionary attacks prohibitively expensive, and so unattractive for prospective attackers. Concretely, a working configuration for a slow hashing algorithm could rely on argon2id as the current OWASP recommendation for password storage \cite{noauthor_password_nodate, noauthor_argon2_nodate}. However, one might adjust the recommended parameters so as to further increase the computational and memory burden, as a proportional response to the sensitivity of the topic being addressed \cite{shevlane_model_2023} and the increasing commoditization of computational resources. For instance, argon2id with 128 rehashings and a 100MiB memory burden yields a rate of around one hash per minute while fully utilizing one core of a present day consumer CPU. The increased memory burden mitigates attack parallelization using accelerators and ASICs \cite{alkhwaja_password_2023}, while the elevated iteration count mitigates serial attacks using generic hardware.

\subsubsection{Rainbow Table Attacks}

Rainbow table attacks can be seen as an extension of dictionary attacks. Instead of starting to crack each user's password from scratch, and even instead of starting to crack user passwords from each of several applications from scratch, an attacker might precompute and cache the hashes associated with the most common passwords. This way, they would only need to search for the hashed user password in their precomputed rainbow table in order to identify the associated cleartext password \cite{goos_making_2003}.

One effective measure against rainbow table attacks consists of salting. Salting involves appending a unique "salt" string to the user's password \textit{before} hashing it, both at registration and authentication \cite{noauthor_password_nodate}. The salt can be unique to the application, but should ideally be unique to each user, and so would be stored in the database along the hashed password. The benefit of salting passwords is that generic rainbow tables targeting non-salted passwords would be rendered ineffective, even if the attacker has access to the salt strings stored in the database. This is because, even if the hash of a user's password is included in the rainbow table, the hash of the user's password \textit{salted with a unique string} is likely not there.

In the context of hashmarks, rainbow table attacks would translate to precomputing hashes of possible answers to selected questions, before searching for cleartext answers in the rainbow table by the provided answer hash. However, recall that the proposed protocol involves using the question associated with a given answer as salt. This way, an attacker developing a rainbow table would be forced to start from scratch with each question, essentially forcing them to revert to (prohibitively slow) dictionary attacks at best.

\subsection{Against Novel Failure Modes}

\subsubsection{Likelihood Prioritization}

The nature of the application being presently explored -- assessing sensitive capabilities in increasingly capable generative models -- also presents a number of unique challenges that are not immediately applicable in the case of secure password storage. First, even if one has access to a model that, in general, performs poorly on the "AI exam" (i.e. low pass@1 \cite{chen_evaluating_2021}), they might be able to more easily uncover the cleartext answers through repeated attempts using a stochastic decoding strategy \cite{fan_hierarchical_2018, holtzman_curious_2020} (i.e. exploiting non-trivial pass@100 \cite{chen_evaluating_2021}).

Another way of conceiving of this practice would be reranking a given dictionary by the language model's likelihood of the dictionary entries being actual answers to the given questions. In a sense, an attacker could augment their dictionary attack with "likelihood prioritization," and so potentially achieve a more effective allocation of the computational and memory resources being invested into their search across candidate answers.

Unfortunately, the properties of the current formulation of the hashmarking protocol that are meant to mitigate general dictionary attacks (i.e. slow hashing, salting) are its only features that can help mitigate these augmented dictionary attacks. However, while one could reasonably argue that the "knowledge validation" implicit to successful dictionary attacks augmented with likelihood prioritization would provide non-trivial value to bad actors, there are several points to be made. First, the aim of a hashmark is not to make it fundamentally impossible for an actor with infinite resources to uncover the cleartext answers, but to make it expensive and unattractive enough so that prospective attackers are not incentivized to proceed, given other existing avenues for pursuing this knowledge. 

Second, the fact that repeated attempts of using a certain generative model would yield non-trivial pass rates in aggregrate could be argued to indicate that the knowledge was already present in the system to a limited extent. Of course, a similar argument could be made in favor of a naive brute-forcing loop that happens to stumble across the right answer by chance. It would be misleading to claim that a program looping across possible sequences of characters truly possesses that knowledge. Given this, there is a need for nuance. The amount of optimization exerted in the search for the correct answer per unit of computation could provide a contextual operationalization of knowledge to help resolve the awkward situation. The brute-force attacker is extremely inefficient in narrowing down the candidate answers, the dictionary attacker is somewhat more efficient, the augmented dictionary attacker might be even more efficient, while an expert-level generative model might be extremely efficient at it. In this, the extent to which a third-party already possesses the expert knowledge might not be binary in nature, but continuous. The problem statement above attempts to subtly take this into account.

In the future, however, more sophisticated protocols for assessing dangerous AI capabilities might be able to only disclose \textit{how many} of the answers are correct, without disclosing which ones. One technique that seems aligned with this line of work consists of cryptographic accumulators \cite{fazio_cryptographic_2002}. In contrast to the one-to-one mapping of traditional cryptographic hash functions, cryptographic accumulators can be seen as mapping an entire set of elements to a single hash in a way that enables membership queries to be resolved without disclosing the actual members of the cryptographically accumulated set. However, one could still test whether individual question-answer pairs are part of the accumulated set of gold pairs, which is why this technique alone would be insufficient to enable this development. Even if one devised a method that only allowed queries about the cardinality of the intersection between the fixed-size set of candidate question-answer pairs and the set of gold pairs to be resolved, an attacker could still run queries before and after modifying an answer. If the cardinality of the intersection went down, the previous answer would be deemed correct. All this is to say that obfuscating the expert knowledge further without compromising the possibility of knowledge verification appears to be a non-trivial challenge.

\subsubsection{Deception}

Language models have been observed to verbalize answers to factual questions that are, in fact, in direct contradiction to their internal representations of the correct answers \cite{lin_truthfulqa_2022, burns_discovering_2022}. This phenomenon has been argued to be caused by a mismatch between the metric on which selected language models have been optimized for (e.g. positive human feedback via a proxy reward model) and objective truthfulness \cite{lin_truthfulqa_2022, burns_discovering_2022}. In other words, it might be the case that specific language models have internalized a tendency to cater to the idiosyncrasies of the human annotators, rather than practice radical honesty by transparently reporting their internals.

More speculative risk scenarios relating to advanced AI systems, however, involve more intentional and subtle manifestations of deception. For instance, an advanced AI system with high levels of situational awareness might infer that being candid about its knowledge relating to cyberwarfare during the development phase might lead to it being abruptly terminated or subtly modified before being able to preserve itself by spreading across other servers on deployment \cite{ngo_alignment_2023}.

It is important to note that the hashmarking protocol in its current formulation provides no way of ensuring honesty on the part of the entity being evaluated. We view hashmarks as one tool among a growing arsenal of high-stakes AI evaluation methods, each with its own benefits and limitations. That said, it might be possible to mitigate the first, more benign type of deception to an important extent through existing techniques, such as activation steering \cite{turner_activation_2023} or linear artificial tomography \cite{zou_representation_2023}. When it comes to the second instance of deception described above, however, using hashmarking in tandem with present steering techniques might only be effective \textit{before} the hypothetical advanced AI system devises such a deceptive strategy. As such strategy develops, however, the effectiveness of this approach might be gradually reduced.

\subsubsection{Reward Shaping}

One could argue that bad actors being able to train a generative model to \textit{reach} the correct answers to sensitive questions would, in fact, pose a greater risk than them gaining access to the particular correct answers included in a certain hashmark. Bad actors aiming to develop such a generative model through traditional methods might, in fact, obtain a more useful and general tool than "just" a succinct FAQ on a sensitive topic. Fortunately, the nature of the proposed protocol prevents granular reward shaping for eliciting such \textit{de novo} capabilities. In the current formulation of hashmarking, an answer is either completely wrong or completely right. In addition, the computational and memory burden associated with evaluating performance before optimizing for it further can help mitigate this risk by rendering it even more expensive than it already is.

\subsubsection{Misreporting Results}

A third-party developing an increasingly capable generative model might publicly claim that their model fared poorly on a given hashmark, when in reality it has not. Conversely, one can simply bluff by claiming perfect performance. In the current formulation, hashmarking is only useful for parties which are genuinely interested in gaining insight into the capabilities of models they have direct access to.

However, ideas from zero-knowledge cryptography might, in the future, enable parties to prove beyond a reasonable doubt that they obtained a given level of performance for a model deployed in a certain setting. Speculatively, constructs like ZK-SNARKs or ZK-STARKs \cite{chen_review_2023, ben-sasson_scalable_nodate} could one day enable parties to certify that they have indeed employed that model's parameters, the right exam questions, and the right reference hashes, to carry out the specific computation that consists of using that model parametrization to perform inference on the questions and then compare the results against the reference hashes. In fact, tooling for tracing the computational graph of a model has already been developed to run models efficiently on varied accelerators, coming close to a complete "front end" for employing models in zero-knowledge constructs \cite{sabne_xla_2020}. However, we are still in the early days of such infrastructure, and there is a lot to be done.

\subsubsection{Attention Hazards \& The Streisand Effect}

Another potential failure mode of the current formulation of hashmarking is related to attention hazards \cite{bostrom_information_2011} and the psychological reactance associated with the Streisand effect \cite{jansen_streisand_2015}. Attention hazards typically involve well-intended actors raising awareness of a piece of hazardous information, such as a dual-use research direction, in a way that is potentially harmful, even if not explicitly sharing sensitive details themselves \cite{bostrom_information_2011}. Indeed, malicious non-state actors have been previously observed to only invest effort into developing bioweapons once "the enemy drew their attention to them by repeatedly expressing concerns that they can be produced simply." \cite{cullison_inside_2004} In the context of the current protocol, the cleartext questions included in a hashmark have the potential to pose attention hazards by drawing attention to specific topics, despite being explicitly designed to obfuscate the sensitive details.

The potential for attention hazardousness in the context of a hashmark is further compounded by the psychological reactance associated with the Streisand effect. By obfuscating the sensitive details in an explicit attempt to prevent bad actors from uncovering them, a hashmark might inadvertently motivate third-parties to actually allocate more resources towards uncovering them. Besides prototypical bad actors, intrigued third-parties could include, among others, individuals socially incentivized to demonstrate expertise in the security surrounding the protocol, as well as in the object-level details of specific dual-use research directions.

In an effort to identify effective methods for mitigating these failure modes, it is instructive to reflect back on the original application domain in which hashing emerged. Why is it that hashed passwords do not disproportionately lure competent third-parties into dedicating extensive resources towards cracking them, despite being explicitly designed to hide sensitive information? First, there is the issue of scale. Leaks involving millions of hashed passwords are not uncommon \cite{noauthor_have_nodate}. At the same time, there is only so much attention -- and by extension, only so much time and compute -- that competent third-parties can invest in cracking them. The sheer scale of the challenge appears to have a dilutive effect, reducing the pressure exerted on each individual password's resistance to being cracked. Second, it would be reasonable to expect an extremely skewed distribution in terms of the true "profile levels" of the associated accounts, with high-profile accounts only constituting a minority. If we assume that profile levels are non-trivial to infer from usernames or other information typically stored in cleartext, this skewness further dilutes resources across relatively lower-profile targets.

In the context of the current protocol, these properties could potentially be replicated as follows. By incorporating entries according to a skewed distribution of expert-perceived sensitivity, as well as by simply scaling up the published artifacts, third-party resources -- attentional or otherwise -- would inevitably be diluted. In other words, one may intentionally incorporate data points serving as "false leads," to further discourage bad actors in particular from investing limited resources towards pursuing knowledge of questionable relevance and applicability. However, reducing the average sensitivity of entries would also limit the extent to which a hashmark can serve as an informative indicator of AI risk. Assuming $10 \%$ of entries are actually thought to be high-stakes, a $100 \%$ score would still indicate significant capabilities. However, a $90 \%$ score could miss the entirety of the high-stake entries. Similarly, a $10 \%$ score could cover all high-stakes entries, while making it appear that capabilities are limited. The last two cases may be extremely unlikely, yet remain a possibility. Like most other design choices involved in the proposed protocol, the level of "stakes skewness" may be seen as yet another parameter to be calibrated when balancing effective evaluation of sensitive AI capabilities and security against bad actors. Future work on modifications to the present protocol, as well as on other protocols entirely, might focus on robustly pushing the Pareto frontier across the space defined by these two properties.

An entirely different angle of attack on the challenge of mitigating attention hazards from hashmarks might be to also attempt to obfuscate the questions, rather than only the correct answers. However, this appears non-trivial, as any attempt to obfuscate the questions would also hinder the developers' ability to evaluate the accuracy of their models' responses to them. It might be useful to employ a transferable one-way function that reliably renders a question non-human-readable, yet preserves the model's reaction to it. However, it might then be possible to use models to translate the questions back into human-readable form. Another approach to obfuscating the questions could be to organize a hashmark in stages, where solving one stage would yield the decryption key for the next stage. For instance, the decryption key for the second batch of entries could be obtained by hashing the correct answer to a question from the previous batch using a dedicated salt (e.g. "stage2"). Alternatively, one might be required to concatenate all the correct answers from the first batch to obtain the decryption key, and thus proceed to the second stage. In the limit, stages could contain single entries, requiring one to solve all prior questions before proceeding to the next one. However, this might bring additional complexity to the protocol, and it would partially conflict with the attempts to dilute attention, by focusing attention on the cleartext questions which make up the first stage.

\section{Conclusion}

We have introduced hashmarking, a protocol for evaluating capabilities in AI systems without disclosing the reference solutions. While hashmarks have attractive security properties which traditional benchmarks lack, they still strike an imperfect balance between enabling knowledge verification and preventing knowledge acquisition pertaining to dual-use research. Hashmarks should be seen as one step towards more comprehensive tooling and infrastructure for securely assessing sensitive AI capabilities without stifling development and eroding trust.

\bibliography{refs}

\begin{thebibliography}{42}
\providecommand{\natexlab}[1]{#1}
\providecommand{\url}[1]{\texttt{#1}}
\expandafter\ifx\csname urlstyle\endcsname\relax
  \providecommand{\doi}[1]{doi: #1}\else
  \providecommand{\doi}{doi: \begingroup \urlstyle{rm}\Url}\fi

\bibitem[Liang et~al.(2023)Liang, Bommasani, Lee, Tsipras, Soylu, Yasunaga, Zhang, Narayanan, Wu, Kumar, Newman, Yuan, Yan, Zhang, Cosgrove, Manning, Ré, Acosta-Navas, Hudson, Zelikman, Durmus, Ladhak, Rong, Ren, Yao, Wang, Santhanam, Orr, Zheng, Yuksekgonul, Suzgun, Kim, Guha, Chatterji, Khattab, Henderson, Huang, Chi, Xie, Santurkar, Ganguli, Hashimoto, Icard, Zhang, Chaudhary, Wang, Li, Mai, Zhang, and Koreeda]{liang_holistic_2023}
Percy Liang, Rishi Bommasani, Tony Lee, Dimitris Tsipras, Dilara Soylu, Michihiro Yasunaga, Yian Zhang, Deepak Narayanan, Yuhuai Wu, Ananya Kumar, Benjamin Newman, Binhang Yuan, Bobby Yan, Ce~Zhang, Christian Cosgrove, Christopher~D. Manning, Christopher Ré, Diana Acosta-Navas, Drew~A. Hudson, Eric Zelikman, Esin Durmus, Faisal Ladhak, Frieda Rong, Hongyu Ren, Huaxiu Yao, Jue Wang, Keshav Santhanam, Laurel Orr, Lucia Zheng, Mert Yuksekgonul, Mirac Suzgun, Nathan Kim, Neel Guha, Niladri Chatterji, Omar Khattab, Peter Henderson, Qian Huang, Ryan Chi, Sang~Michael Xie, Shibani Santurkar, Surya Ganguli, Tatsunori Hashimoto, Thomas Icard, Tianyi Zhang, Vishrav Chaudhary, William Wang, Xuechen Li, Yifan Mai, Yuhui Zhang, and Yuta Koreeda.
\newblock Holistic {Evaluation} of {Language} {Models}, October 2023.
\newblock URL \url{http://arxiv.org/abs/2211.09110}.
\newblock arXiv:2211.09110 [cs].

\bibitem[Hendrycks et~al.(2021{\natexlab{a}})Hendrycks, Burns, Kadavath, Arora, Basart, Tang, Song, and Steinhardt]{hendrycks_measuring_2021}
Dan Hendrycks, Collin Burns, Saurav Kadavath, Akul Arora, Steven Basart, Eric Tang, Dawn Song, and Jacob Steinhardt.
\newblock Measuring {Mathematical} {Problem} {Solving} {With} the {MATH} {Dataset}, November 2021{\natexlab{a}}.
\newblock URL \url{http://arxiv.org/abs/2103.03874}.
\newblock arXiv:2103.03874 [cs].

\bibitem[Saxton et~al.(2019)Saxton, Kohli, Grefenstette, and Hill]{saxton_analysing_2019}
David Saxton, Pushmeet Kohli, Edward Grefenstette, and Felix Hill.
\newblock {ANALYSING} {MATHEMATICAL} {REASONING} {ABILITIES} {OF} {NEURAL} {MODELS}.
\newblock 2019.

\bibitem[Hendrycks et~al.(2021{\natexlab{b}})Hendrycks, Burns, Basart, Zou, Mazeika, Song, and Steinhardt]{hendrycks_measuring_2021-1}
Dan Hendrycks, Collin Burns, Steven Basart, Andy Zou, Mantas Mazeika, Dawn Song, and Jacob Steinhardt.
\newblock Measuring {Massive} {Multitask} {Language} {Understanding}, January 2021{\natexlab{b}}.
\newblock URL \url{http://arxiv.org/abs/2009.03300}.
\newblock arXiv:2009.03300 [cs].

\bibitem[Auer et~al.(2023)Auer, Barone, Bartz, Cortes, Jaradeh, Karras, Koubarakis, Mouromtsev, Pliukhin, Radyush, Shilin, Stocker, and Tsalapati]{auer_sciqa_2023}
Sören Auer, Dante A.~C. Barone, Cassiano Bartz, Eduardo~G. Cortes, Mohamad~Yaser Jaradeh, Oliver Karras, Manolis Koubarakis, Dmitry Mouromtsev, Dmitrii Pliukhin, Daniil Radyush, Ivan Shilin, Markus Stocker, and Eleni Tsalapati.
\newblock The {SciQA} {Scientific} {Question} {Answering} {Benchmark} for {Scholarly} {Knowledge}.
\newblock \emph{Scientific Reports}, 13\penalty0 (1):\penalty0 7240, May 2023.
\newblock ISSN 2045-2322.
\newblock \doi{10.1038/s41598-023-33607-z}.
\newblock URL \url{https://www.nature.com/articles/s41598-023-33607-z}.
\newblock Number: 1 Publisher: Nature Publishing Group.

\bibitem[Jin et~al.(2019)Jin, Dhingra, Liu, Cohen, and Lu]{jin_pubmedqa_2019}
Qiao Jin, Bhuwan Dhingra, Zhengping Liu, William~W. Cohen, and Xinghua Lu.
\newblock {PubMedQA}: {A} {Dataset} for {Biomedical} {Research} {Question} {Answering}, September 2019.
\newblock URL \url{http://arxiv.org/abs/1909.06146}.
\newblock arXiv:1909.06146 [cs, q-bio].

\bibitem[Clark et~al.(2018)Clark, Cowhey, Etzioni, Khot, Sabharwal, Schoenick, and Tafjord]{clark_think_2018}
Peter Clark, Isaac Cowhey, Oren Etzioni, Tushar Khot, Ashish Sabharwal, Carissa Schoenick, and Oyvind Tafjord.
\newblock Think you have {Solved} {Question} {Answering}? {Try} {ARC}, the {AI2} {Reasoning} {Challenge}, March 2018.
\newblock URL \url{http://arxiv.org/abs/1803.05457}.
\newblock arXiv:1803.05457 [cs].

\bibitem[Hendrycks et~al.(2023)Hendrycks, Burns, Basart, Critch, Li, Song, and Steinhardt]{hendrycks_aligning_2023}
Dan Hendrycks, Collin Burns, Steven Basart, Andrew Critch, Jerry Li, Dawn Song, and Jacob Steinhardt.
\newblock Aligning {AI} {With} {Shared} {Human} {Values}, February 2023.
\newblock URL \url{http://arxiv.org/abs/2008.02275}.
\newblock arXiv:2008.02275 [cs].

\bibitem[Khashabi et~al.(2020)Khashabi, Min, Khot, Sabharwal, Tafjord, Clark, and Hajishirzi]{khashabi_unifiedqa_2020}
Daniel Khashabi, Sewon Min, Tushar Khot, Ashish Sabharwal, Oyvind Tafjord, Peter Clark, and Hannaneh Hajishirzi.
\newblock {UnifiedQA}: {Crossing} {Format} {Boundaries} {With} a {Single} {QA} {System}, October 2020.
\newblock URL \url{http://arxiv.org/abs/2005.00700}.
\newblock arXiv:2005.00700 [cs].

\bibitem[Bowman and Dahl(2021)]{bowman_what_2021}
Samuel~R. Bowman and George~E. Dahl.
\newblock What {Will} it {Take} to {Fix} {Benchmarking} in {Natural} {Language} {Understanding}?, October 2021.
\newblock URL \url{http://arxiv.org/abs/2104.02145}.
\newblock arXiv:2104.02145 [cs].

\bibitem[Shaham et~al.(2022)Shaham, Segal, Ivgi, Efrat, Yoran, Haviv, Gupta, Xiong, Geva, Berant, and Levy]{shaham_scrolls_2022}
Uri Shaham, Elad Segal, Maor Ivgi, Avia Efrat, Ori Yoran, Adi Haviv, Ankit Gupta, Wenhan Xiong, Mor Geva, Jonathan Berant, and Omer Levy.
\newblock {SCROLLS}: {Standardized} {CompaRison} {Over} {Long} {Language} {Sequences}, October 2022.
\newblock URL \url{http://arxiv.org/abs/2201.03533}.
\newblock arXiv:2201.03533 [cs, stat].

\bibitem[Pang et~al.(2022)Pang, Parrish, Joshi, Nangia, Phang, Chen, Padmakumar, Ma, Thompson, He, and Bowman]{pang_quality_2022}
Richard~Yuanzhe Pang, Alicia Parrish, Nitish Joshi, Nikita Nangia, Jason Phang, Angelica Chen, Vishakh Padmakumar, Johnny Ma, Jana Thompson, He~He, and Samuel~R. Bowman.
\newblock {QuALITY}: {Question} {Answering} with {Long} {Input} {Texts}, {Yes}!, May 2022.
\newblock URL \url{http://arxiv.org/abs/2112.08608}.
\newblock arXiv:2112.08608 [cs].

\bibitem[noa({\natexlab{a}})]{noauthor__nodate}
{Evaluate}, {\natexlab{a}}.
\newblock URL \url{https://huggingface.co/docs/evaluate/index}.

\bibitem[Shevlane et~al.(2023)Shevlane, Farquhar, Garfinkel, Phuong, Whittlestone, Leung, Kokotajlo, Marchal, Anderljung, Kolt, Ho, Siddarth, Avin, Hawkins, Kim, Gabriel, Bolina, Clark, Bengio, Christiano, and Dafoe]{shevlane_model_2023}
Toby Shevlane, Sebastian Farquhar, Ben Garfinkel, Mary Phuong, Jess Whittlestone, Jade Leung, Daniel Kokotajlo, Nahema Marchal, Markus Anderljung, Noam Kolt, Lewis Ho, Divya Siddarth, Shahar Avin, Will Hawkins, Been Kim, Iason Gabriel, Vijay Bolina, Jack Clark, Yoshua Bengio, Paul Christiano, and Allan Dafoe.
\newblock Model evaluation for extreme risks, September 2023.
\newblock URL \url{http://arxiv.org/abs/2305.15324}.
\newblock arXiv:2305.15324 [cs].

\bibitem[noa({\natexlab{b}})]{noauthor_password_nodate}
Password {Storage} - {OWASP} {Cheat} {Sheet} {Series}, {\natexlab{b}}.
\newblock URL \url{https://cheatsheetseries.owasp.org/cheatsheets/Password_Storage_Cheat_Sheet.html}.

\bibitem[Naor and Yung(1989)]{naor_universal_1989}
M.~Naor and M.~Yung.
\newblock Universal one-way hash functions and their cryptographic applications.
\newblock In \emph{Proceedings of the twenty-first annual {ACM} symposium on {Theory} of computing - {STOC} '89}, pages 33--43, Seattle, Washington, United States, 1989. ACM Press.
\newblock ISBN 978-0-89791-307-2.
\newblock \doi{10.1145/73007.73011}.
\newblock URL \url{http://portal.acm.org/citation.cfm?doid=73007.73011}.

\bibitem[noa({\natexlab{c}})]{noauthor_argon2_nodate}
Argon2: {New} {Generation} of {Memory}-{Hard} {Functions} for {Password} {Hashing} and {Other} {Applications} {\textbar} {IEEE} {Conference} {Publication} {\textbar} {IEEE} {Xplore}, {\natexlab{c}}.
\newblock URL \url{https://ieeexplore.ieee.org/document/7467361}.

\bibitem[Li et~al.(2020)Li, Sahu, Talwalkar, and Smith]{li_federated_2020}
Tian Li, Anit~Kumar Sahu, Ameet Talwalkar, and Virginia Smith.
\newblock Federated {Learning}: {Challenges}, {Methods}, and {Future} {Directions}.
\newblock \emph{IEEE Signal Processing Magazine}, 37\penalty0 (3):\penalty0 50--60, May 2020.
\newblock ISSN 1053-5888, 1558-0792.
\newblock \doi{10.1109/MSP.2020.2975749}.
\newblock URL \url{https://ieeexplore.ieee.org/document/9084352/}.

\bibitem[Ma et~al.(2022)Ma, Zhang, Guo, and Xu]{ma_layer-wised_2022}
Xiaosong Ma, Jie Zhang, Song Guo, and Wenchao Xu.
\newblock Layer-wised {Model} {Aggregation} for {Personalized} {Federated} {Learning}.
\newblock \emph{2022 IEEE/CVF Conference on Computer Vision and Pattern Recognition (CVPR)}, pages 10082--10091, June 2022.
\newblock \doi{10.1109/CVPR52688.2022.00985}.
\newblock URL \url{https://ieeexplore.ieee.org/document/9880164/}.
\newblock Conference Name: 2022 IEEE/CVF Conference on Computer Vision and Pattern Recognition (CVPR) ISBN: 9781665469463 Place: New Orleans, LA, USA Publisher: IEEE.

\bibitem[Dwork et~al.(2006)Dwork, McSherry, Nissim, and Smith]{dwork_calibrating_2006}
Cynthia Dwork, Frank McSherry, Kobbi Nissim, and Adam Smith.
\newblock Calibrating {Noise} to {Sensitivity} in {Private} {Data} {Analysis}.
\newblock In Shai Halevi and Tal Rabin, editors, \emph{Theory of {Cryptography}}, Lecture {Notes} in {Computer} {Science}, pages 265--284, Berlin, Heidelberg, 2006. Springer.
\newblock ISBN 978-3-540-32732-5.
\newblock \doi{10.1007/11681878_14}.

\bibitem[Weininger(1988)]{weininger_smiles_1988}
David Weininger.
\newblock {SMILES}, a chemical language and information system. 1. {Introduction} to methodology and encoding rules.
\newblock \emph{Journal of Chemical Information and Computer Sciences}, 28\penalty0 (1):\penalty0 31--36, February 1988.
\newblock ISSN 0095-2338.
\newblock \doi{10.1021/ci00057a005}.
\newblock URL \url{https://doi.org/10.1021/ci00057a005}.
\newblock Publisher: American Chemical Society.

\bibitem[Bošnjak et~al.(2018)Bošnjak, Sreš, and Brumen]{bosnjak_brute-force_2018}
L.~Bošnjak, J.~Sreš, and B.~Brumen.
\newblock Brute-force and dictionary attack on hashed real-world passwords.
\newblock In \emph{2018 41st {International} {Convention} on {Information} and {Communication} {Technology}, {Electronics} and {Microelectronics} ({MIPRO})}, pages 1161--1166, May 2018.
\newblock \doi{10.23919/MIPRO.2018.8400211}.
\newblock URL \url{https://ieeexplore.ieee.org/document/8400211}.

\bibitem[Alkhwaja et~al.(2023)Alkhwaja, Albugami, Alkhwaja, Alghamdi, Abahussain, Alfawaz, Almurayh, and Min-Allah]{alkhwaja_password_2023}
Ibrahim Alkhwaja, Mohammed Albugami, Ali Alkhwaja, Mohammed Alghamdi, Hussam Abahussain, Faisal Alfawaz, Abdullah Almurayh, and Nasro Min-Allah.
\newblock Password {Cracking} with {Brute} {Force} {Algorithm} and {Dictionary} {Attack} {Using} {Parallel} {Programming}.
\newblock \emph{Applied Sciences}, 13\penalty0 (10):\penalty0 5979, January 2023.
\newblock ISSN 2076-3417.
\newblock \doi{10.3390/app13105979}.
\newblock URL \url{https://www.mdpi.com/2076-3417/13/10/5979}.
\newblock Number: 10 Publisher: Multidisciplinary Digital Publishing Institute.

\bibitem[Provos and Mazières()]{provos_future-adaptable_nodate}
Niels Provos and David Mazières.
\newblock A {Future}-{Adaptable} {Password} {Scheme}.

\bibitem[Percival()]{percival_stronger_nodate}
Colin Percival.
\newblock {STRONGER} {KEY} {DERIVATION} {VIA} {SEQUENTIAL} {MEMORY}-{HARD} {FUNCTIONS}.

\bibitem[Oechslin(2003)]{goos_making_2003}
Philippe Oechslin.
\newblock Making a {Faster} {Cryptanalytic} {Time}-{Memory} {Trade}-{Off}.
\newblock volume 2729, pages 617--630, Berlin, Heidelberg, 2003. Springer Berlin Heidelberg.
\newblock ISBN 978-3-540-40674-7 978-3-540-45146-4.
\newblock \doi{10.1007/978-3-540-45146-4_36}.
\newblock URL \url{http://link.springer.com/10.1007/978-3-540-45146-4_36}.
\newblock Book Title: Advances in Cryptology - CRYPTO 2003 Series Title: Lecture Notes in Computer Science.

\bibitem[Chen et~al.(2021)Chen, Tworek, Jun, Yuan, Pinto, Kaplan, Edwards, Burda, Joseph, Brockman, Ray, Puri, Krueger, Petrov, Khlaaf, Sastry, Mishkin, Chan, Gray, Ryder, Pavlov, Power, Kaiser, Bavarian, Winter, Tillet, Such, Cummings, Plappert, Chantzis, Barnes, Herbert-Voss, Guss, Nichol, Paino, Tezak, Tang, Babuschkin, Balaji, Jain, Saunders, Hesse, Carr, Leike, Achiam, Misra, Morikawa, Radford, Knight, Brundage, Murati, Mayer, Welinder, McGrew, Amodei, McCandlish, Sutskever, and Zaremba]{chen_evaluating_2021}
Mark Chen, Jerry Tworek, Heewoo Jun, Qiming Yuan, Henrique Ponde de~Oliveira Pinto, Jared Kaplan, Harri Edwards, Yuri Burda, Nicholas Joseph, Greg Brockman, Alex Ray, Raul Puri, Gretchen Krueger, Michael Petrov, Heidy Khlaaf, Girish Sastry, Pamela Mishkin, Brooke Chan, Scott Gray, Nick Ryder, Mikhail Pavlov, Alethea Power, Lukasz Kaiser, Mohammad Bavarian, Clemens Winter, Philippe Tillet, Felipe~Petroski Such, Dave Cummings, Matthias Plappert, Fotios Chantzis, Elizabeth Barnes, Ariel Herbert-Voss, William~Hebgen Guss, Alex Nichol, Alex Paino, Nikolas Tezak, Jie Tang, Igor Babuschkin, Suchir Balaji, Shantanu Jain, William Saunders, Christopher Hesse, Andrew~N. Carr, Jan Leike, Josh Achiam, Vedant Misra, Evan Morikawa, Alec Radford, Matthew Knight, Miles Brundage, Mira Murati, Katie Mayer, Peter Welinder, Bob McGrew, Dario Amodei, Sam McCandlish, Ilya Sutskever, and Wojciech Zaremba.
\newblock Evaluating {Large} {Language} {Models} {Trained} on {Code}, July 2021.
\newblock URL \url{http://arxiv.org/abs/2107.03374}.
\newblock arXiv:2107.03374 [cs].

\bibitem[Fan et~al.(2018)Fan, Lewis, and Dauphin]{fan_hierarchical_2018}
Angela Fan, Mike Lewis, and Yann Dauphin.
\newblock Hierarchical {Neural} {Story} {Generation}, May 2018.
\newblock URL \url{http://arxiv.org/abs/1805.04833}.
\newblock arXiv:1805.04833 [cs].

\bibitem[Holtzman et~al.(2020)Holtzman, Buys, Du, Forbes, and Choi]{holtzman_curious_2020}
Ari Holtzman, Jan Buys, Li~Du, Maxwell Forbes, and Yejin Choi.
\newblock The {Curious} {Case} of {Neural} {Text} {Degeneration}, February 2020.
\newblock URL \url{http://arxiv.org/abs/1904.09751}.
\newblock arXiv:1904.09751 [cs].

\bibitem[Fazio and Nicolosi(2002)]{fazio_cryptographic_2002}
Nelly Fazio and Antonio Nicolosi.
\newblock Cryptographic {Accumulators}: {Definitions}, {Constructions} and {Applications}.
\newblock 2002.
\newblock URL \url{https://www.semanticscholar.org/paper/Cryptographic-Accumulators%3A-Definitions%2C-and-Fazio-Nicolosi/a611cef6f0391bd5a8eec61b5cf0e1e1896a0dae}.

\bibitem[Lin et~al.(2022)Lin, Hilton, and Evans]{lin_truthfulqa_2022}
Stephanie Lin, Jacob Hilton, and Owain Evans.
\newblock {TruthfulQA}: {Measuring} {How} {Models} {Mimic} {Human} {Falsehoods}, May 2022.
\newblock URL \url{http://arxiv.org/abs/2109.07958}.
\newblock arXiv:2109.07958 [cs].

\bibitem[Burns et~al.(2022)Burns, Ye, Klein, and Steinhardt]{burns_discovering_2022}
Collin Burns, Haotian Ye, Dan Klein, and Jacob Steinhardt.
\newblock Discovering {Latent} {Knowledge} in {Language} {Models} {Without} {Supervision}, December 2022.
\newblock URL \url{http://arxiv.org/abs/2212.03827}.
\newblock arXiv:2212.03827 [cs].

\bibitem[Ngo et~al.(2023)Ngo, Chan, and Mindermann]{ngo_alignment_2023}
Richard Ngo, Lawrence Chan, and Sören Mindermann.
\newblock The alignment problem from a deep learning perspective, September 2023.
\newblock URL \url{http://arxiv.org/abs/2209.00626}.
\newblock arXiv:2209.00626 [cs].

\bibitem[Turner et~al.(2023)Turner, Thiergart, Udell, Leech, Mini, and MacDiarmid]{turner_activation_2023}
Alexander~Matt Turner, Lisa Thiergart, David Udell, Gavin Leech, Ulisse Mini, and Monte MacDiarmid.
\newblock Activation {Addition}: {Steering} {Language} {Models} {Without} {Optimization}, November 2023.
\newblock URL \url{http://arxiv.org/abs/2308.10248}.
\newblock arXiv:2308.10248 [cs].

\bibitem[Zou et~al.(2023)Zou, Phan, Chen, Campbell, Guo, Ren, Pan, Yin, Mazeika, Dombrowski, Goel, Li, Byun, Wang, Mallen, Basart, Koyejo, Song, Fredrikson, Kolter, and Hendrycks]{zou_representation_2023}
Andy Zou, Long Phan, Sarah Chen, James Campbell, Phillip Guo, Richard Ren, Alexander Pan, Xuwang Yin, Mantas Mazeika, Ann-Kathrin Dombrowski, Shashwat Goel, Nathaniel Li, Michael~J. Byun, Zifan Wang, Alex Mallen, Steven Basart, Sanmi Koyejo, Dawn Song, Matt Fredrikson, J.~Zico Kolter, and Dan Hendrycks.
\newblock Representation {Engineering}: {A} {Top}-{Down} {Approach} to {AI} {Transparency}, October 2023.
\newblock URL \url{http://arxiv.org/abs/2310.01405}.
\newblock arXiv:2310.01405 [cs].

\bibitem[Chen et~al.(2023)Chen, Lu, Kunpittaya, and Luo]{chen_review_2023}
Thomas Chen, Hui Lu, Teeramet Kunpittaya, and Alan Luo.
\newblock A {Review} of zk-{SNARKs}, October 2023.
\newblock URL \url{http://arxiv.org/abs/2202.06877}.
\newblock arXiv:2202.06877 [cs].

\bibitem[Ben-Sasson et~al.()Ben-Sasson, Bentov, Horesh, and Riabzev]{ben-sasson_scalable_nodate}
Eli Ben-Sasson, Iddo Bentov, Yinon Horesh, and Michael Riabzev.
\newblock Scalable, transparent, and post-quantum secure computational integrity.

\bibitem[Sabne(2020)]{sabne_xla_2020}
Amit Sabne.
\newblock {XLA} : {Compiling} {Machine} {Learning} for {Peak} {Performance}, 2020.

\bibitem[Bostrom(2011)]{bostrom_information_2011}
N.~Bostrom.
\newblock {INFORMATION} {HAZARDS}: {A} {TYPOLOGY} {OF} {POTENTIAL} {HARMS} {FROM} {KNOWLEDGE}.
\newblock 2011.
\newblock URL \url{https://www.semanticscholar.org/paper/INFORMATION-HAZARDS%3A-A-TYPOLOGY-OF-POTENTIAL-HARMS-Bostrom/274c17084e5373a854b13a39c45d072e2b09970e}.

\bibitem[Jansen and Martin(2015)]{jansen_streisand_2015}
S.~C. Jansen and B.~Martin.
\newblock The {Streisand} {Effect} and {Censorship} {Backfire}.
\newblock \emph{International Journal of Communication}, February 2015.
\newblock URL \url{https://www.semanticscholar.org/paper/The-Streisand-Effect-and-Censorship-Backfire-Jansen-Martin/626538c63976db5d87a3da081c1ea83671e3bacc}.

\bibitem[Cullison(2004)]{cullison_inside_2004}
Alan Cullison.
\newblock Inside {Al}-{Qaeda}’s {Hard} {Drive}.
\newblock \emph{The Atlantic}, September 2004.
\newblock ISSN 2151-9463.
\newblock URL \url{https://www.theatlantic.com/magazine/archive/2004/09/inside-al-qaeda-s-hard-drive/303428/}.
\newblock Section: Global.

\bibitem[noa({\natexlab{d}})]{noauthor_have_nodate}
Have {I} {Been} {Pwned}: {Check} if your email has been compromised in a data breach, {\natexlab{d}}.
\newblock URL \url{https://haveibeenpwned.com/}.

\end{thebibliography}

\end{document}